\documentclass[11pt]{article}

\usepackage[T1]{fontenc}
\usepackage{lmodern}
\usepackage[a4paper,margin=1in]{geometry}
\usepackage{amsmath}
\usepackage{amssymb}
\usepackage{amsthm}
\usepackage{mathtools}
\usepackage{graphicx}
\usepackage{float}
\usepackage{tabularx}
\usepackage{booktabs}
\usepackage{microtype}
\usepackage[numbers,sort&compress]{natbib}
\usepackage{hyperref}
\hypersetup{hidelinks,pdftitle={Critical Dynamics of AI Self-Improvement},pdfauthor={Mikhail Burtsev}}
\usepackage{tikz}
\usetikzlibrary{arrows.meta,positioning}

\graphicspath{{figures/}}

\newtheorem{proposition}{Proposition}

\newcommand{\RAI}{\mathcal{R}_{\mathrm{AI}}}
\newcommand{\dd}{\mathrm{d}}

\title{Recursive Criticality of AI Self-Improvement}
\author{Mikhail Burtsev \\ \small London Institute for Mathematical Sciences \\ \small mb@lims.ac.uk}
\date{}

\begin{document}

\maketitle

\begin{abstract}

AI is increasingly used in the R\&D process that produces future AI systems. We study the conditions under which this feedback becomes self-amplifying. Our model describes how the rate of AI capability growth depends on baseline research productivity, recursive feedback, and the increasing difficulty of making further research progress. We derive a recursive reproduction number, $\RAI$, that determines whether incremental improvements are amplified or damped across development cycles. This quantity compares the strength of recursive feedback with the rate at which further progress becomes more difficult. When $\RAI>1$, the effects of incremental improvements compound across development cycles, placing the system in a self-amplifying regime. When $\RAI<1$, their effects weaken across cycles. The transition depends on the structure of the AI R\&D feedback loop and need not occur at any particular level of model capability. A system can therefore enter a self-amplifying regime before acceleration becomes visible, while rapid progress can also occur without self-amplification. In the minimal model, higher baseline research productivity can accelerate progress without changing whether the system is self-amplifying. At high research throughput, the duration of the development cycle becomes a limiting timescale for amplification. Increasing research difficulty can subsequently end a period of self-amplification. Extending the model to multiple research actors shows that improvements shared across organizations can make the overall research ecosystem self-amplifying even when no individual actor is. The framework identifies measurable properties of AI R\&D systems that can help distinguish recursive amplification from rapid progress driven by other sources, including the strength of recursive feedback, how effectively improvements propagate into successor systems, development-cycle duration, and the increasing difficulty of further progress. \footnote{The code and computational notebook used to reproduce the numerical results and figures are available at \url{https://github.com/burtsev/recursive-criticality-ai}.}
\end{abstract}

\section{Introduction}
\label{sec:introduction}

Recursive AI self-improvement (RSI) is usually understood as a positive feedback loop in AI capabilities. AI research agents are beginning to perform work that lies inside the process used to build future AI systems. They write and debug code, propose experiments, analyse results, reproduce papers, operate software tools, and increasingly sustain technical work over longer horizons~\cite{Kwa2025,chan2024mle,Wijk2024,Starace2025,toledo2026ai}. RSI is therefore better viewed as a property of an AI--R\&D system than of an isolated, fully autonomous agent. The system may include human researchers, laboratories, compute infrastructure, evaluation procedures and organizations that integrate and deploy successor models.

This paper asks three questions. Under what conditions does AI assistance to AI R\&D cross from ordinary acceleration into a self-amplifying regime? How do feedback delay and a hardening research progress determine whether that regime produces a small transient displacement or significantly compressed transition to artificial general intelligence (AGI) and artificial super intelligence (ASI)? How are these dynamics changed by physical resource limits, competition and collaboration between research actors?

We use the term \emph{recursive criticality} for the condition at which an incremental improvement in AI research capability generates enough additional future R\&D productivity to outweigh the increasing difficulty of making further progress. This balance is defined by a \textit{recursive reproduction number} $\RAI$. When  $\RAI>1$, incremental gains amplify across development cycles. When $\RAI>1$, their effects weaken. This yields the following criterion:

\vspace{6pt}
\noindent\textbf{Onset of self-amplifying RSI.}
\textit{The transition to self-amplifying recursive improvement occurs when the AI system crosses a critical stability boundary, not when it reaches a particular level of intelligence. This boundary is crossed when $\RAI$ rises above unity, so that gains in research capability reproduce across development cycles faster than the effective research frontier hardens.}
\vspace{6pt}

The possibility of a machine-driven intelligence explosion dates to Good~\cite{Good1965}, and subsequent work developed the conceptual foundations of superintelligence and self-modifying agents \cite{Hutter2012,Schmidhuber2006,Everitt2016}. More recent economic models place AI directly inside the production of future technological progress, treating it as an input to research whose effects depend on automation, bottlenecks, diminishing returns and the productivity of complementary resources \cite{Aghion2017,Davidson2023,Besiroglu2022,JonesAI2025,Davidson2026}. Related models examine resource-constrained growth and distinguish bounded self-refinement from open-ended research loops \cite{Jafari2025,Chen2026}. Cunningham et al.~\cite{Cunningham2026} derive a closely related condition for self-sustaining acceleration in terms of economic elasticities, while recent analyses of the transition from AGI to ASI emphasize that scaling, paradigm change, recursive improvement and collective AI systems may operate simultaneously rather than as separate pathways \cite{Genewein2026}.

At the same time, empirical work is making the underlying feedback process increasingly measurable. General benchmarks provide increasingly systematic comparisons of model capability across tasks and generations, while human-calibrated task horizons estimate the duration and complexity of work that AI agents can complete reliably \cite{Hardt2026,HoBenchmarks2025,Kwa2025}. Research-oriented evaluations such as MLE-bench, RE-Bench and PaperBench move closer to the relevant setting by testing agents on extended machine-learning engineering, experimentation and research-replication tasks \cite{chan2024mle,Wijk2024,Starace2025,toledo2026ai}. These evaluations increasingly measure components of the feedback loop, but they do not yet identify how an improvement in AI research capability causally affects the productivity of subsequent AI development. Proposed measures of AI-R\&D automation extend this perspective to the organizational level by tracking adoption, researcher time and the effect of AI assistance on subsequent scientific and engineering progress \cite{Chan2026}.

Taken together, existing work brings many of the ingredients of recursive improvement into view, but not yet the dynamics that determine whether the resulting feedback is stable. In particular, it leaves open how the transition to self-amplifying improvement depends on the delay between research and deployment, the progressive exhaustion or hardening of research opportunities, and the coupling between multiple research actors.

We develop a minimal model in which AI capability evolves through baseline research productivity, delayed recursive gain and a capability-dependent research frontier. We then extend the framework to include physical deployment constraints and research networks through which improvements propagate between laboratories, firms and nations. The resulting model separates interventions that primarily alter the pace of development from those that change the recursive dynamics themselves or reshape the network through which recursive gains spread. The numerical scenarios are conditional demonstrations and not probabilistic forecasts. They show how alternative assumptions generate qualitatively different trajectories and identify the measurements that could distinguish among them.

We make four principal contributions.

First, we formulate recursive AI self-improvement as a local stability transition in an AI-enabled R\&D system. The resulting recursive reproduction number,
\[
\RAI=\frac{\chi a}{\sigma},
\]
compares realized recursive gain with the local hardening of the research frontier. Incremental capability gains amplify across development cycles when $\RAI>1$ and are damped when $\RAI<1$. In the minimal model, the critical boundary does not depend on baseline research throughput, so rapid capability growth and recursive self-amplification are distinct phenomena. A system can become supercritical before the resulting acceleration becomes visible.

Second, we characterize how the duration and intensity of recursive amplification depend on feedback delay and the available research frontier. At high research throughput, the successor-development cycle becomes a limiting timescale for amplification. As research opportunities become harder to exploit, increasing frontier hardness can return a supercritical system to a subcritical regime. Recursive amplification can therefore be strong but transient within a fixed research paradigm.

Third, we separate the recursive regime from both research scale and physical deployment constraints. Greater compute, expenditure, or researcher effort can accelerate capability growth without changing recursive criticality, while changes in recursive gain, operational closure, feedback delay, or transfer between actors alter the feedback dynamics themselves. Physical infrastructure can additionally constrain deployable capability without directly changing the recursive regime.

Fourth, we extend the framework to coupled research actors. Recursive criticality is then determined by the spectral radius of a reproduction matrix that captures both within-actor feedback and cross-actor transfer. A research ecosystem can therefore become supercritical even when every actor is individually subcritical. This formulation also identifies quantities that can be estimated empirically, including recursive gain, operational closure, development-cycle duration, frontier hardening, and the transfer of improvements across organizations.

\section{A dynamical model of recursive self-improvement}
\label{sec:model}

\subsection{Recursive loop in AI development}
\label{sec:system}

We model recursive self-improvement at the level of an AI--R\&D system rather than an isolated AI agent. The system includes the AI models used for research together with the processes required to generate, evaluate, train and deploy successor systems. Human researchers may remain inside this system boundary.

\begin{figure}[H]
\centering
\begin{tikzpicture}[
    node distance=2.6cm,
    box/.style={
        draw,
        rounded corners,
        align=center,
        minimum width=3.0cm,
        minimum height=1.05cm
    },
    >=Latex
]
\node[box] (cap) {AI capability\\$x(t)$};
\node[box, right=of cap] (rnd)
    {AI--R\&D\\productivity};
\node[box, right=of rnd] (succ)
    {validated\\successor};
\node[box, above=0.4cm of rnd] (res)
    {baseline research\\$r(t)$};
\draw[->]
    (res) -- (rnd);
\draw[->]
    (cap) -- node[above]{gain $a$} (rnd);
\draw[->]
    (rnd) -- node[above]{closure $\chi$} node[below]{delay $\tau$} (succ);
\draw[->]
    (succ.south west)
    to[out=-150,in=-25]
    node[below]{further capability}
    (cap.south east);
\draw[->, loop below, looseness=5]
    (cap) to
    node[below]{frontier hardening $\sigma$}
    (cap);

\end{tikzpicture}

\caption{\textbf{Minimal AI self-improvement loop.} Current AI capability affects subsequent research productivity. A fraction of this potential gain survives the development pipeline and returns after an end-to-end delay as capability in a validated successor system. Frontier hardening reduces the productivity of further improvement as capability advances.}
\label{fig:rsi-loop}
\end{figure}
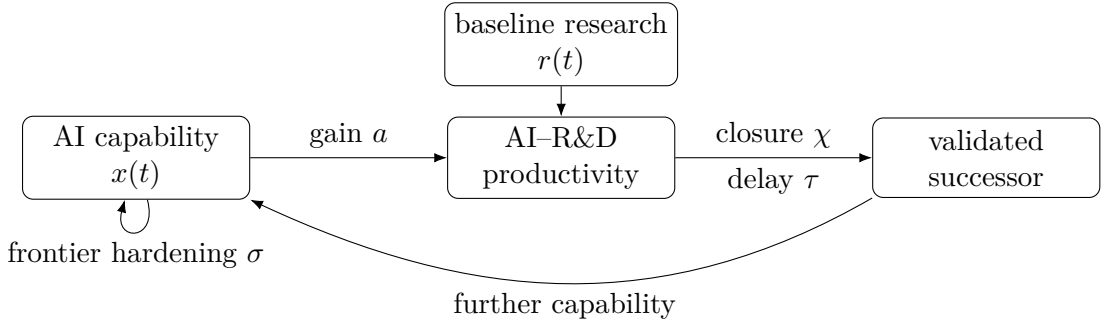

Let $x(t)$ denote an AI capability. The model, illustrated in Fig.~\ref{fig:rsi-loop}, separates four elements of the development process.

\begin{enumerate}

\item Baseline progress is governed by an exogenous \textit{research-productivity} scale $r(t)$.

\item Existing AI capability can increase the productivity of subsequent AI development. We represent this effect by \textit{recursive gain} $a(x,t)$ and an end-to-end \textit{delay} $\tau(t)$ between an improvement in capability and its return as greater research productivity.

\item Only part of the potential recursive gain may survive the full development pipeline. We denote this transmitted fraction by the \textit{operational closure} $\chi(x,t)$.

\item Progress becomes increasingly difficult as capability approaches the research frontier defined by local\textit{ hardening rate}  $\sigma(x)$.

\end{enumerate}

To connect the abstract state variable $x$ to observable research performance, let $c_t(q,\epsilon;\mathbf p)$ denote the minimum resource cost of completing task $q$ at tolerated loss $\epsilon$, evaluated at a fixed vector of resource prices $\mathbf p$. For a stable panel of tasks $\mathcal Q$ with weights $\phi(q)$, define
\begin{equation}
x(t)
=
\int_{\mathcal Q}
\phi(q)
\ln
\frac{
c_{t_0}(q,\epsilon;\mathbf p)
}{
c_t(q,\epsilon;\mathbf p)
}
\,\dd\mu(q).
\label{eq:capability-coordinate}
\end{equation}
Thus $x$ measures resource-adjusted improvement at fixed task composition and performance. It is a local coordinate for research capability, not a universal scalar measure of intelligence.

Let $a(x,t)$ denote the recursive gain from stronger AI research capability to subsequent research productivity when downstream use is unconstrained. Let $0\leq\chi(x,t)\leq1$ denote the fraction transmitted through the research pipeline. The realized recursive gain is
\begin{equation}
g(x,t)=\chi(x,t)a(x,t).
\label{eq:realized-gain}
\end{equation}

Operational closure $\chi$ is a property of the development process, not simply of model autonomy. A system can have high closure while retaining substantial human participation if AI-generated research contributions are evaluated and integrated efficiently. Conversely, a highly autonomous workflow can have low closure when its outputs fail evaluation, cannot be executed safely, or do not propagate into successor systems. Human review, experimental capacity, secure execution, and other bottlenecks can bind as throughput rises, so $\chi$ may depend on operating scale as well as capability.

Frontier hardness $\sigma(x,t)$ measures the decline in direct improvement productivity as capability advances with recursive transfer held fixed. The delay $\tau$ is the end-to-end interval before an improvement returns as increased research capability. Together, $g$, $\sigma$, and $\tau$ determine the local stability and manifestation rate of recursive feedback.

\subsection{A delayed model of recursive criticality}
\label{sec:model}

We model recursive AI self-improvement by combining baseline research productivity, frontier difficulty and delayed recursive feedback:
\begin{equation}
\boxed{
\dot{x}(t)=r(t)f[x(t)] e^{\Phi(x[t-\tau(t)],t)}.
}
\label{eq:nonlinear-model}
\end{equation}
Here $r(t)>0$ sets the baseline scale of research progress, $f(x)>0$ describes how the productivity of direct improvement changes as the research frontier advances. The exponential term represents the multiplicative effect of recursive feedback. In the absence of recursive amplification, $\Phi=0$ and progress reduces to the
baseline term $r(t)f(x)$.

We define the feedback potential $\Phi$ so that its local derivative equals the realized recursive gain,
\begin{equation}
\frac{\partial \Phi(x,t)}{\partial x}
=
g(x,t)
=
\chi(x,t)a(x,t).
\label{eq:gain-potential}
\end{equation}
Thus $g(x,t)$ measures how strongly an incremental increase in research capability
raises the productivity returned through the recursive loop.

We quantify frontier hardening by the rate at which direct improvement productivity declines with capability,
\begin{equation}
\boxed{
\sigma(x)
=
-
\frac{\dd \ln f(x)}{\dd x}.
}
\label{eq:frontier-hardness}
\end{equation}
Positive $\sigma$ means that advancing the frontier reduce the productivity of subsequent improvement. 

Consider a reference trajectory $\bar{x}(t)$ during an interval in which the coefficients change slowly relative to the perturbation dynamics. Set $v=\dot{\bar{x}}(t_0)>0$, $g=g[\bar{x}(t_0),t_0]$, $\sigma=\sigma[\bar{x}(t_0)]$, and $\tau=\tau(t_0)$. A small perturbation $x=\bar{x}+\xi$ then obeys
\begin{equation}
\dot{\xi}(t)=-v\sigma\,\xi(t)+vg\,\xi(t-\tau),
\label{eq:linearized}
\end{equation}
with characteristic equation
\begin{equation}
\lambda+v\sigma=vg e^{-\lambda\tau}.
\label{eq:characteristic}
\end{equation}
The local balance becomes transparent after defining
\begin{equation}
\boxed{\RAI=\frac{g}{\sigma}=\frac{\chi a}{\sigma}.}
\label{eq:rai}
\end{equation}

\begin{proposition}[Local recursive criticality]
Suppose that $v>0$, $\sigma>0$, $g\geq0$ and $\tau\geq0$. All characteristic roots of Eq.~\eqref{eq:linearized} have negative real part for $\RAI<1$, $\lambda=0$ is a characteristic root at $\RAI=1$, and a positive real characteristic root exists for $\RAI>1$.
\end{proposition}

This result is a scalar instance of stability theory for positive delay systems \cite{Briat2017}. Cunningham et al.~\cite{Cunningham2026} independently obtain a closely related unit-threshold condition for self-sustaining acceleration from recursive elasticities. In the present model, the threshold marks a stability boundary for perturbations around a delayed capability trajectory.  Locally, $\mathcal{R}_{\mathrm{AI}}>1$ implies amplification of small recursive perturbations around the reference trajectory, but does not by itself imply indefinite growth of total capability.

The mechanism behind the threshold is simple. Recursive gain must offset replace the marginal loss of research productivity created by the hardening frontier before a capability perturbation can reproduce. Baseline throughput $v$ controls how much development occurs per unit time, but in the minimal model it scales both the damping and the feedback terms. Greater compute, expenditure or researcher effort can therefore accelerate capability growth without necessarily changing whether the system is subcritical or supercritical.

Feedback delay instead determines how rapidly the local regime becomes visible. Near the critical point, the dominant characteristic root is
\[
\lambda_*
\simeq
\frac{
v\sigma
\left(
\mathcal{R}_{\mathrm{AI}}-1
\right)
}{
1+
v\sigma\mathcal{R}_{\mathrm{AI}}\tau
}.
\]
The dominant root passes smoothly through zero, which means that a newly supercritical process may initially be difficult to distinguish from ordinary acceleration. Its amplification rate becomes more apparent as research throughput rises or as the successor-development cycle becomes shorter. 

As research throughput grows, the delay itself becomes limiting. For fixed $\RAI>1$ and $\tau>0$,
\[
\lambda_*
\longrightarrow
\frac{
\ln\mathcal{R}_{\mathrm{AI}}
}{
\tau
}
\qquad
\text{as}
\qquad
v\sigma\tau\longrightarrow\infty.
\]
Increasing research throughput can therefore accelerate an amplifying mode, but cannot make it arbitrarily fast without shortening the interval through which improvements return in a successor system. Other limits on effective research throughput, including bounded parallelizability and coordination, can bind before this delay-limited regime is reached \cite{Trammell2026}. 

Rapid capability growth is therefore neither necessary nor sufficient evidence of recursive criticality. A high baseline research rate, greater compute availability or substantial human effort can produce fast subcritical progress, while a supercritical system can initially amplify too slowly to be distinguished from an ordinary acceleration. Diagnosing the regime therefore requires estimates of recursive gain and frontier hardening across successive development cycles, rather than the slope of a benchmark trajectory considered in isolation.

\subsection{Finite recursive runway}
\label{sec:finite}

The local criticality condition determines whether capability perturbations amplify at the current state, but it does not require the same recursive mechanism to remain supercritical as capability advances. A simple finite-frontier model is
\begin{equation}
 f(x)=\left(1-\frac{x}{X}\right)^\beta,
 \qquad 0\leq x<X,
 \quad \beta>0,
 \label{eq:frontier}
\end{equation}
which gives
\begin{equation}
\sigma(x)
=
\frac{\beta}{X-x}.
\label{eq:finite-frontier-hardness}
\end{equation}
The frontier becomes progressively harder as $x$ approaches the effective limit $X$.

\begin{proposition}[Finite-frontier termination]
Suppose that $x(t)\to X<\infty$,
$\sigma[x(t)]\to\infty$, and $g[x(t)]$ remains bounded. Then
\begin{equation*}
\RAI[x(t)] = \frac{g[x(t)]}{\sigma[x(t)]} \longrightarrow 0,
\end{equation*}
so the system is locally subcritical sufficiently close to the effective frontier.
\end{proposition}

This result makes the duration of a supercritical regime depend on the available recursive runway. Within a fixed research paradigm, the system may cross into a supercritical region, amplify rapidly for some interval, and return to a subcritical regime once frontier hardening overtakes realized recursive gain. A major architectural or scientific advance may extend the frontier $X$, change its shape through $\beta$, or increase the realized gain $g$, so the result should not be read as a claim about a final ceiling. It instead shows that transient supercriticality is compatible with a finite research opportunity set.

\subsection{Coupled research-agent ecosystems}
\label{sec:network}

AI development does not occur inside a single closed loop. Research agents operate within organizations that observe competitors, exchange scientific information, reuse software, recruit from a common labour market, depend on shared infrastructure, and sometimes obtain access to the same models or tools. These channels allow an improvement produced at one actor to change the future research productivity of another. Once such transfer becomes comparable to within-actor recursive gain, the relevant stability question concerns the network as a whole.

Let $\xi_i(t)$ be a perturbation to the research capability of actor $i$. A local multi-actor approximation is
\begin{equation}
\dot{\xi}_i(t)=-v_i\sigma_i\xi_i(t)
+v_i\sum_{j=1}^{n}g_{ij}\xi_j(t-\tau_{ij}),
\label{eq:network}
\end{equation}
where $g_{ii}$ denotes gain generated within actor $i$ and $g_{ij}$ denotes directed transfer from actor $j$ to actor $i$. Normalizing each row by the receiving actor's frontier hardness defines the reproduction matrix
\begin{equation}
K_{ij}=\frac{g_{ij}}{\sigma_i},
\qquad
\mathcal{R}_{\mathrm{net}}=\rho(\mathbf K),
\label{eq:strategic-global-criticality}
\end{equation}
where $\rho(\mathbf K)$ is the spectral radius of the non-negative matrix $\mathbf K$.

\begin{proposition}[Collective criticality]
For the zero-delay system associated with Eq.~\eqref{eq:network}, the locally stable regime satisfies $\rho(\mathbf K)<1$, while the system becomes unstable when $\rho(\mathbf K)>1$. Hence $K_{ii}<1$ for every actor is not sufficient for network stability when cross-actor transfer is strong enough.
\end{proposition}

This result extends recursive criticality from individual development loops to the research ecosystem as a whole. Every actor may remain individually subcritical while the coupled system becomes supercritical through the circulation and recombination of improvements across organizations. An advance originating in one laboratory may improve the tools available to another, whose subsequent work raises research productivity elsewhere before some of those gains diffuse back to the original actor. The combined reproduction of capability across these pathways can exceed the damping imposed by individual research frontiers even though no actor has a within-organization reproduction number above one.

\section{Reference scenarios}
\label{sec:results}

\subsection{Reference no-RSI parameterization}

The numerical experiments are designed to compare dynamical mechanisms rather than to forecast calendar dates. We normalize current capability to $x_0=0$ and the effective frontier of the modeled research paradigm to $X=1$. These values define the coordinate system and do not imply that current AI has zero absolute capability or that $X=1$ represents a fundamental limit.

We represent AGI and ASI by externally specified thresholds $x_{\rm AGI}$ and $x_{\rm ASI}$ on the capability coordinate, with $x_{\rm AGI}<x_{\rm ASI}$. The lower threshold can represent broad competence across a specified set of economically or scientifically relevant tasks, while the upper threshold can represent substantially greater performance under the same evaluation conditions. For any threshold $y$, define the first-crossing time as $T(y) = \inf\{t\geq0:x(t)\geq y\}$, $T_{\rm AGI}=T(x_{\rm AGI})$, $T_{\rm ASI}=T(x_{\rm ASI})$.

This construction separates the placement of AGI and ASI on a capability scale from the dynamics that carry the system between them. Our results concern the latter, conditional on the former. For the reference scenarios we set
\begin{equation*}
x_{\rm AGI}=0.50, \qquad x_{\rm ASI}=0.80. 
\end{equation*} 
These values are illustrative. They place AGI midway between the reference state and the effective frontier, while ASI lies substantially closer to the frontier. The resulting separation is sufficient to distinguish gradual progress, transient recursive amplification and compression of the AGI-to-ASI interval within a common capability coordinate.

We use the reference hardness value $\beta=2$ in the frontier function~\eqref{eq:frontier} as a structural assumption.  This choice makes direct improvement productivity decline quadratically with the remaining headroom while preserving an analytic baseline solution. Under the normalization $X=1$, the corresponding frontier hardness is
\begin{equation*} 
\sigma(x) = \frac{\beta}{X-x} = \frac{2}{1-x}. 
\end{equation*} 
Hardness therefore rises gradually at low capability and diverges as the effective frontier is approached.

To construct the reference trajectory, we remove recursive feedback by setting $\Phi(x,t)=0$, omit additional physical throughput constraints, and hold baseline research productivity fixed at $r(t)=r_{\rm ref}$. The dynamics reduce to
\begin{equation*}
\dot{x}^{(0)}(t) = r_{\rm ref} \left[ 1-x^{(0)}(t) \right]^2, \qquad x^{(0)}(0)=0,
\end{equation*} 
with solution
\begin{equation*} 
x^{(0)}(t) = \frac{r_{\rm ref}t} {1+r_{\rm ref}t}, 
 \qquad 
T^{(0)}(x) = \frac{x} {r_{\rm ref}(1-x)}. 
\end{equation*} 
For the reference threshold positions: 
$T_{\rm AGI}^{(0)} = \frac{1}{r_{\rm ref}}$ and $T_{\rm ASI}^{(0)} = \frac{4}{r_{\rm ref}}$. 
Thus $r_{\rm ref}$ sets the no-RSI AGI and ASI timescales. The single parameter $r_{\rm ref}$ therefore sets the timescale of the no-RSI trajectory.

We choose this timescale to place the reference AGI crossing broadly within the range of recent expert elicitation. The 2023 Expert Survey on Progress in AI, covering 2,778 authors from major AI venues, reported an aggregate 50\% date of 2047 for high-level machine intelligence~\cite{Grace2025}. A later wave of the Longitudinal Expert AI Panel reported a conditional median AGI date of 2050 among 205 experts, with 25\% and 75\% dates of 2039 and 2065~\cite{leap2025}. These surveys use different definitions and exhibit substantial uncertainty, so we use them only to set an illustrative timescale rather than as a probabilistic calibration. Taking $t=0$ to correspond to 2026 and setting the reference AGI crossing to 2050 gives a 24-year baseline timescale
\begin{equation}
\boxed{
r_{\rm ref} = \frac{1}{24} \simeq 0.0417~\mathrm{yr}^{-1}.
}
\label{eq:reference-baseline-rate}
\end{equation}

The capability normalization, frontier shape and baseline rate together fully specify the reference trajectory without recursive feedback. 

\subsection{Reference parameterization of the RSI part}

Existing empirical evidence shows substantial algorithmic progress and improving AI performance on increasingly demanding technical tasks, but it does not separately identify the recursive gain $a$, the operational-closure function $\chi(x)$ or the relevant frontier hardness \cite{Ho2024,Gundlach2025,METRExpenditureHorizon2026}. We therefore treat these quantities as structural scenario parameters. Throughout the reference recursive scenarios, we fix the feedback delay at $\tau=0.5~\mathrm{yr}$ as a plausible interval between a capability improvement and its effect on subsequent AI R\&D productivity.

To isolate the central mechanism of the model, we construct a small family of recursive scenarios in which the only varying parameter is the recursive gain $a$. The operational-closure trajectory, the frontier-hardening exponent and the feedback delay are held fixed across the primary scenarios. Differences in the simulated trajectories can therefore be attributed directly to differences in the strength of recursive AI-to-AI-R\&D feedback.

We model operational closure as a smooth increasing function of capability,
\begin{equation}
\chi(x)
=
\frac{1}
{1+e^{-kx}}.
\label{eq:sigmoid-closure}
\end{equation}
The steepness parameter $k>0$ determines how rapidly the AI--R\&D loop approaches closure as capability increases, with larger values concentrating the transition within a narrower capability interval. Because present capability is normalized to $x_0=0$, Eq.~\eqref{eq:sigmoid-closure} gives $\chi(0)=0.5$. Closure then rises toward unity, while remaining strictly below one at any finite capability. The parameter $a$ therefore represents the recursive gain under ideal operational closure.

For the finite-frontier model with $X=1$, the local recursive reproduction number is
\begin{equation*}
\mathcal{R}_{\rm AI}(x) = \frac{a\,\chi(x)(1-x)}{\beta}.
\label{eq:reference-recursive-reproduction-number}
\end{equation*}
Operational closure rises with capability, while remaining frontier headroom $1-x$ declines. Their product can therefore reach an interior maximum. We define the peak recursive criticality of a parameter combination by
\begin{equation*}
\mathcal{R}_{\rm peak}(a,\beta,k)
=
\max_{0\leq x<1}
\mathcal{R}_{\rm AI}(x).
\label{eq:peak-recursive-criticality}
\end{equation*}

We choose four reference values of $a$ to represent distinct dynamical regimes: subcritical \textit{smooth scaling}, \textit{weak supercritical}, \textit{transient takeoff} and \textit{rapid transition}.

\begin{figure}[t]
\centering
\includegraphics[width=\textwidth]{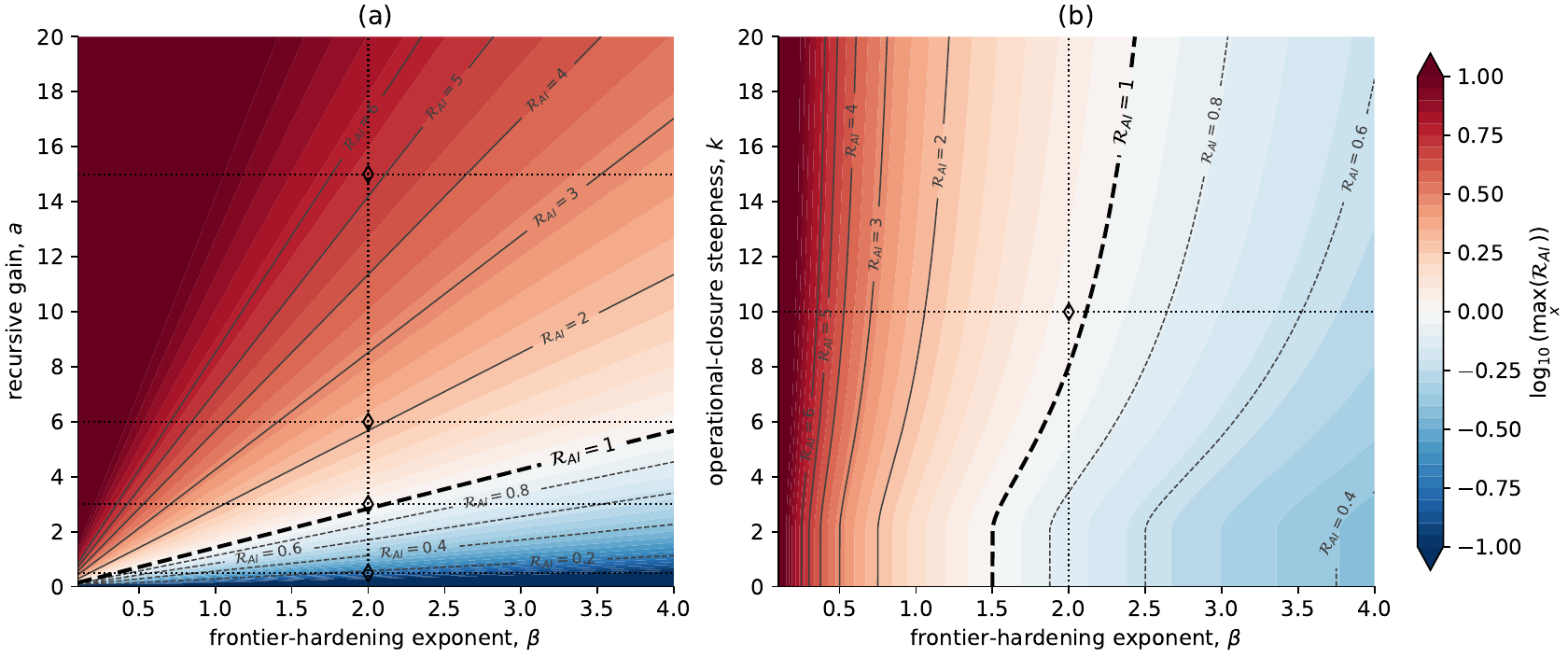}
\caption{
\textbf{Recursive criticality across parameter space.}
\textbf{a}, Peak recursive criticality $\mathcal{R}_{\rm peak}$ as a function of the frontier-hardening exponent $\beta$ and recursive gain $a$, with $k=10$. Diamonds mark the four reference scenarios: subcritical \textit{smooth scaling}, \textit{weak supercritical}, \textit{transient takeoff}, \textit{rapid transition} at $\beta=2$ and
$a\in\{0.5,3,6,15\}$.
\textbf{b}, Dependence of peak criticality on $\beta$ and the operational-closure steepness $k$, with $a=3$. The diamond marks the reference choice $(\beta,k)=(2,10)$. The boundary $\mathcal{R}_{\rm peak}=1$ separates parameter combinations that remain
subcritical from those that enter a locally supercritical regime.
}
\label{fig:reference-criticality-maps}
\end{figure}

The reference value $k=10$ produces a relatively concentrated, but still smooth, increase in operational closure. At $a=3$ and $\beta=2$, corresponding to the \textit{weak-supercritical} scenario, this choice places the system only modestly above the peak critical boundary. We hold $k$ fixed across all primary scenarios considered below.

\subsection{Reference dynamical regimes}
\label{sec:reference-regimes}

\begin{figure}[t]
    \centering
    \includegraphics[width=\textwidth]{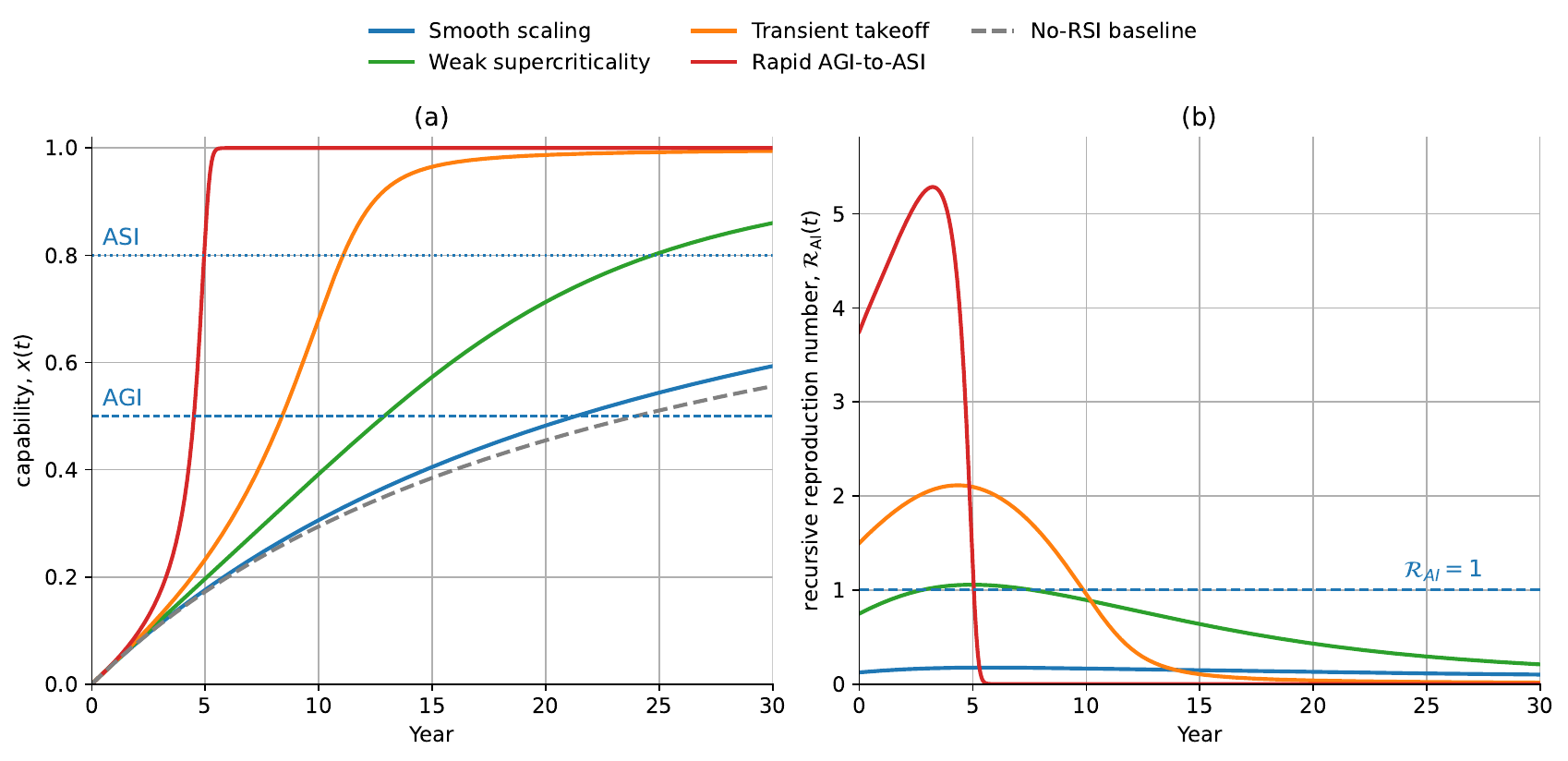}
\caption{
\textbf{Stronger recursive feedback progressively compresses the transition from AGI to ASI, while frontier hardening ultimately suppress supercriticality.}
\textbf{a}, Capability trajectories for recursive gains $a\in\{0.5,3,6,15\}$ and the matched no-RSI baseline. Horizontal lines mark the illustrative AGI and ASI thresholds at $x_{\rm AGI}=0.50$ and $x_{\rm ASI}=0.80$.
\textbf{b}, Corresponding recursive reproduction numbers $\mathcal{R}_{\rm AI}(t)$. The dashed horizontal line marks the local critical boundary $\mathcal{R}_{\rm AI}=1$.
All scenarios use $x_0=0$, $X=1$, $\beta=2$, $k=10$, $\tau=0.5$ yr and $r_{\rm ref}=1/24~{\rm yr}^{-1}$.
}
    \label{fig:reference-scenarios}
\end{figure}

The phase maps in Fig.~\ref{fig:reference-criticality-maps} identify parameter combinations that contain a locally supercritical region. They do not show how long the trajectory remains in that region, how much capability is accumulated there, or where the supercritical episode falls relative to the AGI and ASI thresholds. Figure~\ref{fig:reference-scenarios} follows the four reference parameterizations through time.

Recursive feedback can materially alter the capability trajectory even when the system never becomes supercritical. In the \textit{smooth scaling} scenario, $\mathcal{R}_{\rm AI}(t)<1$ throughout (Fig.~\ref{fig:reference-scenarios}b), yet recursive feedback advances both threshold crossings relative to the no-RSI baseline. Subcriticality therefore rules out self-amplification of local perturbations, but not a substantial cumulative contribution from recursive feedback.

The \textit{weak supercritical} scenario shows why crossing the critical boundary need not coincide with an obvious takeoff. The reproduction number rises only modestly above unity and falls below the boundary well before AGI is reached. Nevertheless, the temporary period of amplification leaves the system on a persistently advanced capability trajectory. Frontier hardening subsequently suppresses further amplification, but it does not undo the capability accumulated during the supercritical episode.

As recursive gain increases, a larger part of the AGI-to-ASI transition occurs while the system is supercritical. In the \textit{rapid transition} scenario, both thresholds are crossed during the same amplification episode and the interval between them becomes very short. The reproduction number then declines rapidly as the trajectory approaches the fixed frontier, where increasing hardness eventually dominates recursive gain. Stronger recursive feedback can therefore change both the timing and concentration of progress without changing the asymptotic capability limit imposed by a fixed $X$.

Table~\ref{tab:reference-transition-times} quantifies these differences. The effect of recursive feedback becomes increasingly pronounced for thresholds farther from the initial state. Relative to the no-RSI trajectory, the subcritical \textit{smooth scaling} scenario advances AGI by about $2.7$ years and ASI by about $22$ years. At the other extreme, the \textit{rapid transition} scenario advances AGI by about $19.5$ years and ASI by about $91$ years, while compressing the AGI-to-ASI interval from $72$ years to less than half a year.

The asymmetry arises because recursive feedback compounds over the capability interval. Its effect on the later ASI threshold can therefore be much larger than its effect on the first AGI crossing. Relatively modest differences in AGI timing can coexist with very large differences in the duration of the subsequent transition.

Taken together, the reference trajectories show that the same initial capability can support markedly different development paths. Subcritical feedback can produce a meaningful acceleration, a brief supercritical episode can leave a durable capability lead, and strong recursive amplification can compress the AGI-to-ASI transition while remaining self-limiting. The trajectory therefore depends on the evolving balance between recursive gain, operational closure and frontier hardening, not on capability level alone.

\begin{table}[t]
\centering
\small
\setlength{\tabcolsep}{4.5pt}
\caption{
Threshold-crossing times for the four reference scenarios and the matched no-RSI baseline. The AGI-to-ASI interval is
$\Delta T_{\rm AGI\rightarrow ASI}=T_{\rm ASI}-T_{\rm AGI}$. \\
}
\label{tab:reference-transition-times}
\begin{tabular}{@{}lrrrr@{}}
\toprule
Scenario
& $a$
& $T_{\rm AGI}$ (yr)
& $T_{\rm ASI}$ (yr)
& $\Delta T_{\rm AGI\rightarrow ASI}$ (yr)
\\
\midrule
No-RSI baseline
& 0
& 24.00
& 96.00
& 72.00
\\
Smooth scaling
& 0.5
& 21.35
& 74.13
& 52.79
\\
Weak supercriticality
& 3
& 12.90
& 24.73
& 11.83
\\
Transient takeoff
& 6
& 8.40
& 11.08
& 2.69
\\
Rapid AGI-to-ASI
& 15
& 4.50
& 4.95
& 0.45
\\
\bottomrule
\end{tabular}
\end{table}

In reference dynamical regimes the effective frontier $X$ is fixed and frontier hardening dominates when capability advances faster than the effective frontier moves. However, $X$ can change during AI development. New model architectures, training methods or theoretical insights can induce a sufficiently rapid outward movement of $X(t)$ and prolong the supercritical regime.

A discrete scientific or algorithmic breakthrough can instead produce a jump $X\longrightarrow X+\Delta X$. The resulting increase in headroom raises $\mathcal{R}_{\mathrm{AI}}$ immediately in the reduced model. A trajectory can therefore undergo several separated recursive episodes as successive paradigms create new opportunities and each opportunity set is later exhausted. The finite-frontier result should accordingly be interpreted as local to a research paradigm rather than as evidence for a single final capability ceiling.

It should be emphasized, that the reference regimes are intended to expose the qualitative consequences of alternative recursive-feedback assumptions rather than to provide probabilistic forecasts. We calibrate $r_{\rm ref}$ to recent expert expectations for AGI timing and choose $\tau$ to represent a plausible AI-R\&D cycle, although both remain uncertain. The frontier-hardening exponent $\beta$, recursive gain $a$ and operational-closure steepness $k$ are not presently constrained by direct empirical estimates and are selected to span distinct dynamical regimes. The periods reported in Fig.~\ref{fig:reference-scenarios}, Table~\ref{tab:reference-transition-times} and for other scenarios should therefore be interpreted only as conditional outputs of the stated parameterization.


\section{Strategic scenarios}
\label{sec:strategy}

\subsection{Strategic competition and network organization}
\label{sec:strategic-competition}

The coupled-agent framework in Sec.~\ref{sec:network} allows strategic conditions to affect AI development through several distinct channels. Competition can mobilize additional investment, compute and research effort, while also changing how much of the AI--R\&D process is delegated to advanced models, how quickly successor systems are developed, and how readily improvements propagate between research actors \cite{GrossmanShapiro1987,Armstrong2016,BuenoDeMesquita2026}. These effects act on different components of the recursive dynamics and need not move together.

We represent changes in baseline research intensity by an effort multiplier $m_r$. Increasing $m_r$ raises the rate at which capability advances, but in the minimal model does not by itself change the local recursive threshold. Changes in recursive gain, operational closure or frontier hardening alter the actor-level reproduction number, while cross-actor transfer changes the network reproduction number $\rho(\mathbf K)$. Strategic competition can consequently accelerate development, alter recursive criticality, and reshape the network through which improvements accumulate and spread.

\begin{figure}[]
\centering
\includegraphics[width=\textwidth]{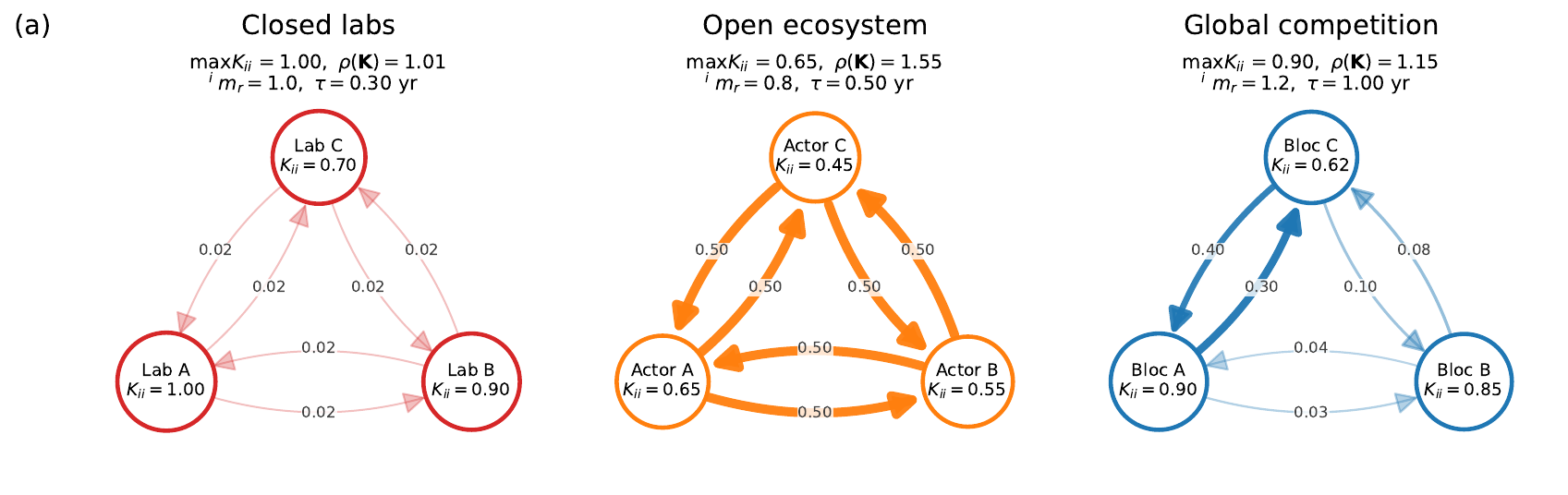}\\
\includegraphics[width=\textwidth]{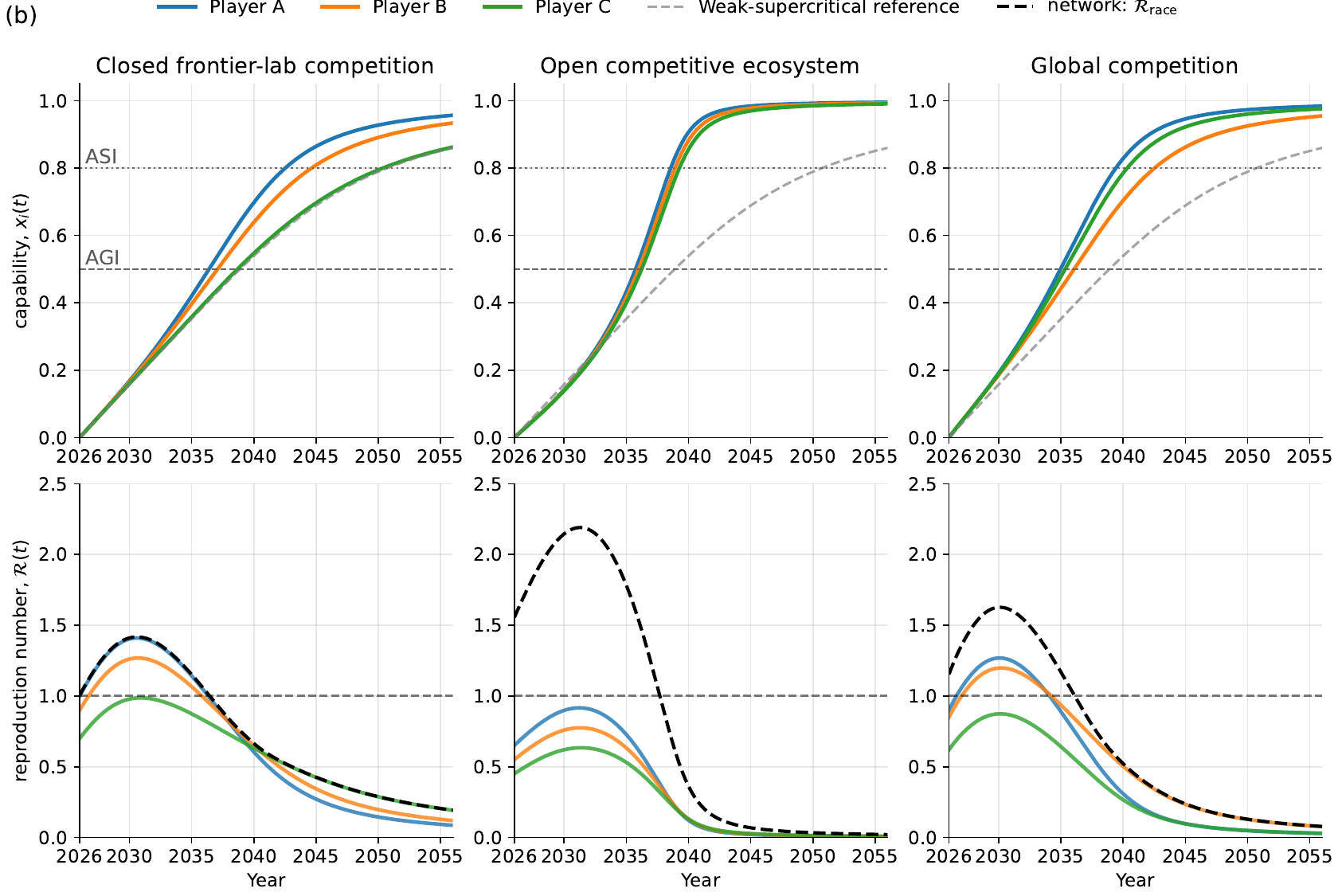}
\caption{
\textbf{Strategic organization changes both the timing and structure of recursive AI development.}
\textbf{a}, Three illustrative AI-development networks, nodes show within-actor reproduction numbers $K_{ii}$ and directed links show cross-actor transfer $K_{ij}$. The spectral radius $\rho(\mathbf{K})$ gives the corresponding network reproduction number at the reference state. The effort multiplier $m_r$ rescales baseline research productivity, while $\tau$ is the characteristic feedback delay. \textit{Closed laboratories} combine strong internal recursive loops with weak exchange. The \textit{open ecosystem} contains individually subcritical actors connected by strong transfer. \textit{Global competition} combines elevated research effort with asymmetric cross-bloc spillovers and a longer feedback delay.
\textbf{b}, Nonlinear capability trajectories (top) and recursive reproduction numbers (bottom) for the same configurations, with the strategic network active from 2026. Coloured curves show individual actors, the grey dashed curve shows the weak-supercritical reference trajectory, and the black dashed curve shows the time-dependent network reproduction number $\mathcal{R}_{\mathrm{net}}(t)$. All strategic parameters are illustrative.
}
\label{fig:strategic-networks}
\end{figure}

We use the weak supercritical reference trajectory from Sec.~\ref{sec:reference-regimes} as a common starting point and compare three illustrative research environments (Fig.~\ref{fig:strategic-networks}a). Each configuration specifies  $m_r$, $\tau$, and a matrix $\mathbf K$ describing within-actor recursive gain and directed transfer between actors. The diagonal elements $K_{ii}$ describe normalized recursive reproduction within actor $i$, while the off-diagonal elements $K_{ij}$ describe the contribution of actor $j$ to the future research productivity of actor $i$. Their spectral radius gives the corresponding network-level recursive criticality, as established in Sec.~\ref{sec:network}.

The reference trajectory defines $m_r=1$, so values above or below unity represent greater or lower effective research throughput relative to the same baseline. The three configurations are chosen to separate the effects of strong internal recursion, broad cross-actor diffusion and elevated competitive effort rather than to represent forecasts of particular firms or nations. Figure~\ref{fig:strategic-networks} compares their network structure and resulting capability dynamics, while Table~\ref{tab:strategic-thresholds} summarizes the corresponding threshold-crossing
times.

The \textit{closed laboratory} scenario concentrates recursive gain within individual organizations. The leading laboratory begins at the local critical boundary, $K_{AA}=1.00$, while the other two remain subcritical. Cross-laboratory transfer is weak, so coupling raises the network reproduction number only modestly. We keep research effort at the reference level, $m_r=1$, and use a short feedback delay of $\tau=0.30$ yr to represent rapid internal development cycles. Capability trajectories then diverge gradually as differences in within-laboratory recursive gain accumulate
over successive cycles.

The \textit{open ecosystem} produces amplification through a different mechanism. All three actors are individually subcritical, but strong off-diagonal transfer allows improvements generated by one actor to raise the subsequent research productivity of the others. Recursive feedback is therefore distributed across the network rather than concentrated inside a single organization. We set the effort multiplier below the reference value, $m_r=0.8$, with a characteristic delay of $\tau=0.50$ yr. Despite the lower baseline research intensity, network coupling produces the shortest AGI-to-ASI transition of the three scenarios, approximately $2.8$ years.

The \textit{global competition} scenario combines higher research effort with a more weakly and asymmetrically connected network. Each bloc is individually subcritical at the reference state, but transfer between the leading blocs raises the network reproduction number to $\rho(\mathbf K)\simeq1.15$. We set $m_r=1.2$ to represent additional resources mobilized by strategic competition and use a longer delay of $\tau=1.0$ yr to represent slower transmission, replication and integration of advances across competing actors. This scenario reaches AGI first, after approximately $9.0$ years, but its AGI-to-ASI interval remains about $4.4$ years, longer than in the open ecosystem despite its greater research effort.

These scenarios are designed to separate mechanisms rather than to estimate the behaviour of particular institutions or geopolitical systems. Greater openness can increase the diffusion of useful advances, while organizational separation can keep recursive gain concentrated within individual actors. Competition can increase the overall rate of research without producing a comparable increase in recursive coupling. The resulting trajectories depend jointly on the reproduction matrix $\mathbf K$, the research-effort multiplier $m_r$, and the feedback delay $\tau$. Research-network structure should therefore be treated as part of the dynamical state of advanced AI development. Aggregate investment provides information about the speed of progress, but it does not determine proximity to recursive criticality. 

\begin{table}[]
\centering
\caption{
\textbf{Threshold-crossing times for the three strategic configurations.} $T_{\mathrm{AGI}}$ and $T_{\mathrm{ASI}}$ are the first crossing times among actors in each network. The final column gives the interval between the first AGI and ASI crossings.\\
}
\label{tab:strategic-thresholds}
\begin{tabular}{lccc}
\hline
\textbf{Configuration}
&
$\boldsymbol{T_{\mathrm{AGI}}}$ \textbf{(yr)}
&
$\boldsymbol{T_{\mathrm{ASI}}}$ \textbf{(yr)}
&
$\boldsymbol{\Delta T_{\mathrm{AGI}\rightarrow\mathrm{ASI}}}$ \textbf{(yr)}
\\
\hline
Closed frontier-lab competition
& 10.40 & 16.53 & 6.13 \\
Open competitive ecosystem
& 9.75 & 12.58 & 2.82 \\
Global competition
& 8.97 & 13.42 & 4.45 \\
\hline
\end{tabular}
\end{table}

\subsection{Hardware and power limits}
\label{sec:hardware-power}

The reference trajectories assume that physical infrastructure can expand as quickly as software-side capability demands. This may fail well before fundamental computational limits become relevant, because frontier systems also require power, cooling, networking and data-centre infrastructure that expand on physical construction timescales \cite{IEAEnergyAI2025}.

Let $C_{\mathrm{phys}}(t)$ denote the effective compute capacity supported by available infrastructure and $C_{\mathrm{req}}(x,t)$ the compute required to instantiate capability
$x$. Deployability requires
\begin{equation}
C_{\mathrm{req}}[x(t),t]
\leq
C_{\mathrm{phys}}(t).
\label{eq:physical-feasibility}
\end{equation}
Algorithmic progress can reduce $C_{\mathrm{req}}$ at fixed capability \cite{Ho2024,Gundlach2025}, so this quantity represents the net compute requirement under the assumptions of each scenario.

We approximate the requirement locally by
\begin{equation}
C_{\mathrm{req}}(x)
=
C_{\mathrm{req}}(x_{\mathrm{AGI}})
e^{\kappa(x-x_{\mathrm{AGI}})},
\label{eq:compute-requirement}
\end{equation}
and define physical headroom at the unconstrained AGI crossing as
\begin{equation}
F_{\mathrm{AGI}}^{\mathrm{soft}}
=
\frac{
C_{\mathrm{phys}}(T_{\mathrm{AGI}}^{\mathrm{soft}})
}{
C_{\mathrm{req}}(x_{\mathrm{AGI}})
}.
\label{eq:physical-headroom-soft-agi}
\end{equation}
The corresponding compute increase between the illustrative AGI and ASI thresholds is
\begin{equation}
Q
=
\frac{
C_{\mathrm{req}}(x_{\mathrm{ASI}})
}{
C_{\mathrm{req}}(x_{\mathrm{AGI}})
}
=
e^{\kappa(x_{\mathrm{ASI}}-x_{\mathrm{AGI}})}.
\label{eq:agi-asi-compute-ratio}
\end{equation}

\begin{figure}[t]
\centering
\includegraphics[width=\textwidth]{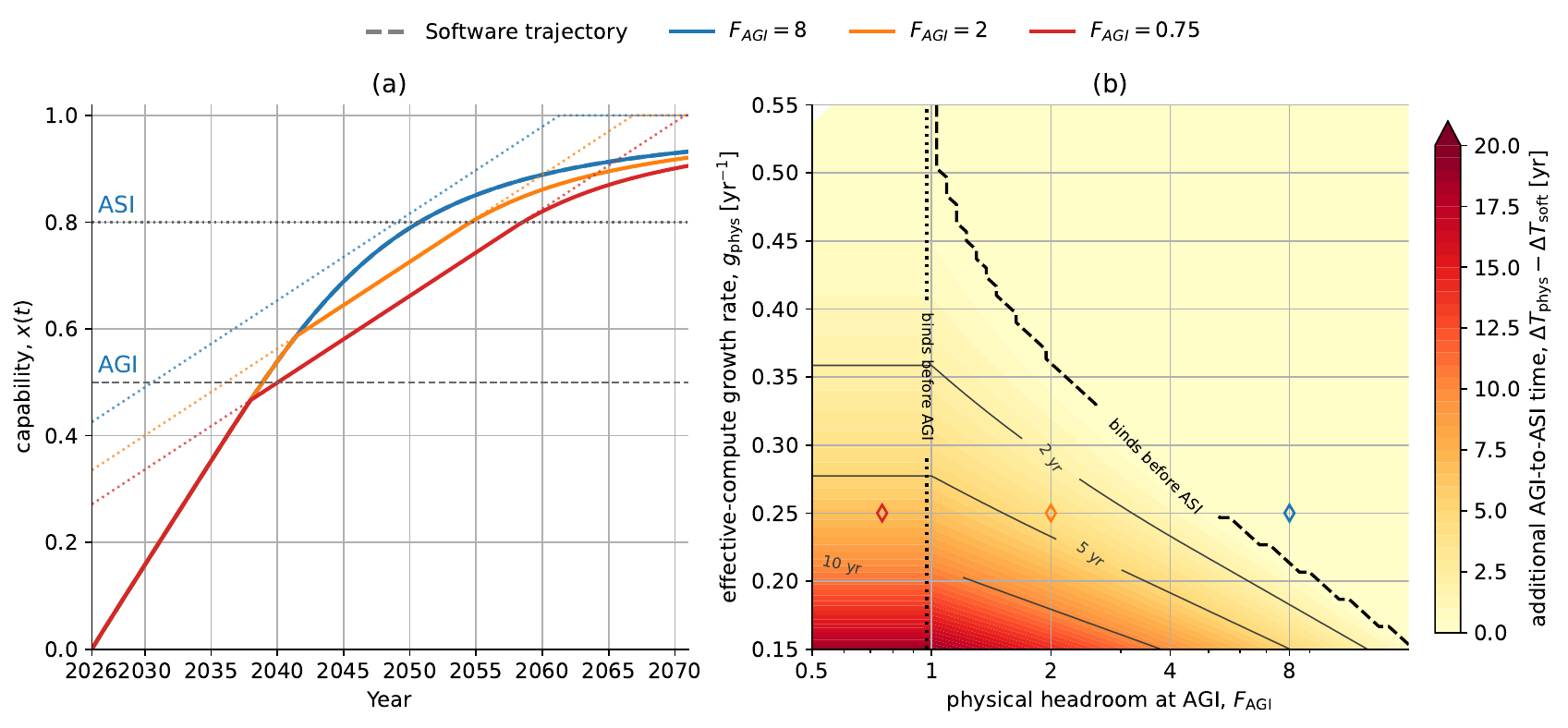}
\caption{\textbf{Physical headroom can separate software-side capability from deployable capability.} \textbf{a}, Weak supercritical reference trajectory for $F_{\mathrm{AGI}}^{\mathrm{soft}}\in\{8,2,0.75\}$. \textbf{b}, Additional AGI-to-ASI time as a function of physical headroom and effective compute growth. Dotted and dashed boundaries mark where the physical constraint first binds before AGI and before ASI. Diamonds correspond to panel~\textbf{a}.}
\label{fig:physical-headroom}
\end{figure}

For illustration, we set $Q=100$ and let effective physical compute grow at $g_{\mathrm{phys}}=0.25~\mathrm{yr}^{-1}$. These values are scenario assumptions rather than estimates of the resource requirements of AGI or ASI. Figure~\ref{fig:physical-headroom} shows three regimes. With $F_{\mathrm{AGI}}^{\mathrm{soft}}=8$, infrastructure remains ahead of the software trajectory. At $F_{\mathrm{AGI}}^{\mathrm{soft}}=2$, capacity becomes limiting between AGI and ASI and lengthens the transition. At $F_{\mathrm{AGI}}^{\mathrm{soft}}=0.75$, the constraint already binds before the unconstrained AGI crossing and delays both thresholds.

Physical compute has a different role from recursive gain. As a capacity constraint, it limits which software-side capabilities can be deployed without directly changing $\mathcal{R}_{\mathrm{AI}}$ at a given capability. Compute can also alter recursive dynamics when it increases operational closure, strengthens research gain or shortens development cycles \cite{WhitfillWu2025}. The scenarios here isolate the first effect.

\section{Conclusion}
\label{sec:conclusion}

Recursive self-improvement is best understood as a dynamical property of an AI-enabled R\&D system. In our model, the transition to self-amplifying improvement occurs when realized recursive gain exceeds the local hardening of the research frontier, so that $\RAI=\chi /  a\sigma>1$. Incremental improvements then amplify across successive development cycles. This transition is determined by the structure of the development process and need not coincide with a particular capability threshold such as AGI. Crossing $\RAI=1$ is a local condition for amplification and does not by itself imply indefinite acceleration or unbounded capability growth.

The framework separates the onset, speed, and persistence of recursive amplification.$\RAI=1$ determines whether incremental gains amplify, while development-cycle delay constrains how rapidly that amplification unfolds. Increasing research difficulty can subsequently return a supercritical system to a subcritical regime, so strong recursive amplification can be transient within a fixed research paradigm. Higher research throughput can produce rapid progress without changing the recursive regime, and a newly supercritical system can initially resemble ordinary acceleration. Recursive gain also depends on operational closure, thus, AI-generated research affects future capability only to the extent that it propagates through evaluation, integration, training, and deployment into successor systems.

These distinctions extend beyond a single research actor. In coupled research ecosystems, improvements can propagate between organizations strongly enough to make the network supercritical even when every actor is individually subcritical. Physical infrastructure introduces a different constraint by limiting which software-side capabilities can be deployed without necessarily changing recursive criticality itself. The numerical scenarios illustrate how these mechanisms can generate qualitatively different development trajectories under alternative assumptions.

The main empirical challenge is therefore to measure the feedback mechanisms directly. Relevant quantities include the causal effect of AI research capability on subsequent R\&D productivity, the fraction of potential gains that propagate into successor systems, development-cycle duration, resource-normalized research productivity, frontier hardening, and the transfer of improvements across actors. Capability growth alone is not sufficient to diagnose recursive criticality. The more informative signal is whether each increment of AI research capability is becoming increasingly effective at producing the next one. If recursive self-improvement emerges, changes in this reproduction of research capability may become detectable before the most visible phase of acceleration.

\bibliographystyle{unsrt}
\bibliography{critical_dynamics}

\end{document}